# Negative Tree Reweighted Belief Propagation

**Qiang Liu**
Department of Computer Science
University of California, Irvine
Irvine, CA, 92697
qliu1@ics.uci.edu

**Alexander Ihler**
Department of Computer Science
University of California, Irvine
Irvine, CA, 92697
ihler@ics.uci.edu

## Abstract

We introduce a new class of lower bounds on the log partition function of a Markov random field which makes use of a reversed Jensen's inequality. In particular, our method approximates the intractable distribution using a linear combination of spanning trees with negative weights. This technique is a lower-bound counterpart to the tree-reweighted belief propagation algorithm, which uses a convex combination of spanning trees with positive weights to provide corresponding upper bounds. We develop algorithms to optimize and tighten the lower bounds over the non-convex set of valid parameter values. Our algorithm generalizes mean field approaches (including naïve and structured mean field approximations), which it includes as a limiting case.

## 1 INTRODUCTION

Probabilistic graphical models provide powerful tools for representing and reasoning about uncertainty. A common difficulty in many learning and inference tasks is calculating the log partition function, or normalization constant of the distribution. This task corresponds to computing the probability of evidence in Bayes nets, and is an often critical component of learning a distribution from observed data. For tree-structured graphical models (i.e., those without cycles), the partition function can be calculated efficiently by variable elimination. However, for graphical models with cycles this calculation is exponential in the tree width of the graph, and is thus often computationally infeasible (Wainwright and Jordan, 2008).

Variational methods for approximate inference provide approximations, often bounds, of the log partition function. Of these, perhaps the most popular is loopy belief propagation (LBP), which provides exact results for tree-structured graphical models and often results in excellent approximations in graphs with cycles. The fixed points of LBP yield stationary points of the Bethe free energy approximation to the log partition function (Yedidia et al., 2004); however, this approximation has few guarantees on quality or whether it under- or over-estimates the partition function, except in certain special cases (Sudderth et al., 2008).

In contrast, the mean field approach provides a guaranteed lower bound on the log partition function. Lower bounds are very attractive in learning algorithms (such as expectation-maximization) which attempt to maximize the data likelihood, since optimizing a lower bound on likelihood provides some assurance of the learned model's quality (Zhang, 1992; Saul and Jordan, 1999; Wainwright and Jordan, 2008). Naïve mean field approximates the original model using a fully factored distribution, while structured mean field (Saul and Jordan, 1995) generalizes this approach to include some dependence among variables.

In the other direction, the tree-reweighted belief propagation (TRBP) algorithm provides a class of upper bounds on the log partition function by exploiting convexity (Wainwright et al., 2005). The viewpoint of convex duality can be used to unify many of these approaches as optimizations of exact or approximate entropies on subsets (lower bounds) or supersets (upper bounds) of the set of valid marginal distributions, or marginal polytope (Wainwright and Jordan, 2008). However, TRBP is generally considered separate and comparatively unrelated to mean field approaches. The two methods have very different properties; for example, TRBP is a convex optimization, while mean field is in general non-convex.

In this paper, we unify and extend these perspectives by proposing a class of lower bounds of the log partition function using a reversed Jensen's inequality. This results in a non-convex optimization problem which provides a lower bound on the log parti-tion function, but whose structure and analysis are closely connected to TRBP. Moreover, this framework encompasses mean field and structured mean field approaches as special, limiting cases, and enables optimizing the approximating structure during inference.

The paper is organized as follows. Section 2 provides notation and briefly reviews tree reweighted belief propagation. In section 3 we establish a reverse Jensen's inequality and propose our lower bound on the log partition function using negative weights. Sections 4–5 derive a counterpart message passing algorithm for our lower bound, then propose an algorithm to improve the bound by optimizing its weights in section 6. We describe connections to mean field and other related work in section 7. Finally, we illustrate the method experimentally in section 8, and conclude with section 9.

## 2 BACKGROUND

Let $G = (V, E)$ be an undirected graph, where $V$ is the set of nodes and $E$ is the set of edges. We define an undirected graphical model over $G$ by placing at each node $i \in V$ a random variable $x_i$ taking values in the discrete space $\mathcal{X}_i = \{0, 1, \cdots, m_i - 1\}$, and let $x = \{x_i | i \in V\}$ be a random vector in the product space $\mathcal{X}^N = \mathcal{X}_1 \times \mathcal{X}_2 \cdots \mathcal{X}_n$. The exponential representation of the graphical model is written in terms of the parameter vector $\theta$ and sufficient statistics $\phi(x)$:

$$p(x|\theta) = \exp\left\{\sum_{\alpha \in \mathcal{I}} \theta_\alpha \phi_\alpha(x) - \Phi(\theta)\right\},$$

$$\Phi(\theta) = \log \sum_{x \in \mathcal{X}^n} \exp\left\{\sum_{\alpha \in \mathcal{I}} \theta_\alpha \phi_\alpha(x)\right\},$$

where $\mathcal{I}$ is the index set of the parameters and sufficient statistics. The quantity $\Phi(\theta)$ is the *log partition function*, which serves to normalize the distribution. A well known property of $\Phi(\theta)$ is its convexity. However, calculation of $\Phi(\theta)$ requires summing over an exponential number of terms, and is a major challenge for learning and inference tasks in graphical models.

For simplicity of presentation, here we consider only pairwise MRFs. We follow Wainwright et al. (2005) and use the *overcomplete* exponential representation, where the statistics are defined as indicator functions:

$$\phi_{i;s}(x) = \delta(x_i = s) \quad , \quad \phi_{ij;st}(x) = \delta(x_i = s, x_j = t)$$

and $\delta$ is the Kroneker delta function, equal to one iff the argument equalities hold. Then, the index set is the union of $\mathcal{I}(V) = \{(i, s) | i \in V, s \in \mathcal{X}\}$ and edge indices $\mathcal{I}(E) = \{(ij, st) | (i, j) \in E, s, t \in \mathcal{X}\}$. We also define $\theta_i(x_i) = \theta_{i;x_i}$ and $\theta_{ij}(x_i, x_j) = \theta_{ij;x_i,x_j}$ for convenience.

### 2.1 TREE REWEIGHTED BP

Wainwright et al. (2005) introduces a class of upper bound on $\Phi(\theta)$ that arise immediately from the property of convexity. The idea is to split the parameter vector into a linear combination of parameter vectors of tractable distributions (such as spanning trees), and then apply Jensen's inequality.

Let $\mathcal{T} = \{T_r\}_{r=1}^{R}$, with $T_r = (V, E_r)$ be a collection of spanning trees of $G$. In principle, we could include arbitrary subtrees $T_r$, but spanning trees are sufficient to provide optimal bounds (Wainwright et al., 2005). However, we will relax $\mathcal{T}$ to include arbitrary subtrees, including the fully disconnected subgraph, when we draw our connections to mean field.

We decompose the exponential parameter vector $\theta$ into a collection $\bar{\theta} = \{\theta^r\}_{r=1}^{R}$, such that $\sum_r \theta^r = \theta$ and for each $r$, $\theta_r$ is an exponential parameter vector that respects the structure of $T_r$. More formally, this requires that $\theta_{ij}^r = 0$ if $(i, j) \notin E_r$. For convenience, we let $T(i, j) = \{r | (i, j) \in E_r\}$ be the collection of spanning trees that include edge $(i, j)$.

Suppose we also have a set of weights $w = [w_1, \cdots, w_R]$ over the spanning trees, such that $\sum_r w_r = 1$. Following the convention of Wainwright et al. (2005), we call $\mu_{ij} = \sum_{r \in T(i,j)} w_r$ the edge appearance probability of edge $(i, j)$, although in the sequel we will not require that $w_r \geq 0$ and thus $\mu_{ij}$ is not interpretable as an actual "probability" in our framework.

If $w_r \geq 0$, we apply Jensen's inequality:

$$\Phi(\theta) = \Phi\left(\sum_r w_r \frac{\theta^r}{w_r}\right) \leq \sum_r w_r \Phi\left(\frac{\theta^r}{w_r}\right) \triangleq \Psi(\bar{\theta}, w).$$

A natural question is then how to choose $\theta^r$ and $w_r$ to obtain the tightest upper bound. This question is divided into two parts in Wainwright et al. (2005):

*For fixed $w$*, define the optimal bound by

$$\Psi_w^+ \triangleq \arg\min_{\bar{\theta}} \Psi(\bar{\theta}, w) \qquad \text{s.t.} \quad \theta = \sum_r \theta^r.$$

This is a convex optimization problem with linear constraints. Although the number of possible spanning trees is extremely large, this difficulty can be sidestepped using Lagrangian duality, by observing that there are only a few constraints. Wainwright et al. (2005) shows that this dual optimization can be interpreted as minimizing a free energy $\mathcal{F}_{TRBP}$, which is closely related to the Bethe free energy. A simple "message passing" fixed-point algorithm is developed, similar to that of loopy belief propagation.

*To optimize $w$*, it is shown that the free energy $\mathcal{F}_{TRBP}$ depends only on the edge appearance probabilities $\mu_{ij}$.

A conditional gradient method can be used to optimize the $\mu_{ij}$; this is equivalent to a maximum-weight spanning tree problem, in which the mutual information between nodes $i$ and $j$ serves as the weight on edge $(i, j)$. An advantage of TRBP is that one can optimize the parameters without explicitly keeping track of the $w_r$ and $\theta^r$ for each spanning tree.

## 3 REVERSING JENSEN'S

Jensen's inequality yields an upper bound of the convex function $\Phi(\theta)$. Here, we establish a reversed Jensen's inequality, which instead gives a lower bound:

**Lemma 3.1.** *Suppose $f(\theta)$ is a convex function on $\mathbb{R}^m$. Let $\theta_i \in \mathbb{R}^m$, $i = 1, 2, \ldots, n$ be points in $\mathbb{R}^m$, and $w = [w_1, \cdots, w_n]$ be a set of weights satisfying $\sum_{i=1}^n w_i = 1$. If all weights $w_i$ except one are negative, e.g., $w_1 > 1$, $w_i < 0$ for $i = 2, \ldots, n$, then*

$$\sum_{i=1}^n w_i f(\theta_i) \leq f(\sum_{i=1}^n w_i \theta_i). \quad (1)$$

*Sketch of proof.* Let $w_1 > 0$ and define $a = \sum_i w_i \theta_i$ and $b = \frac{1}{1-w_1} \sum_{i>1} w_i \theta_i$, then note $\theta_1 = (a + (w_1 - 1)b)/w_1$, apply Jensen's inequality to $\theta_1$, then again to $b$, and rearrange terms. □

A one-dimensional version of this inequality and some extensions are discussed in Mitrinovic et al. (1993). If more than one weight $w_i$ is positive, the inequality does not hold in general. In fact, we show in Section 7 that loopy belief propagation can be interpreted as corresponding to multiple positive and some negative weights, giving an interpretation as to why loopy BP provides neither an upper nor lower bound of the log partition function in general.

With the reversed Jensen's inequality, it is straightforward to get a lower bound of the log partition function:

$$\Phi(\theta) = \Phi\left(\sum_r w_r \frac{\theta^r}{w_r}\right) \geq \sum_r w_r \Phi\left(\frac{\theta^r}{w_r}\right) \triangleq \Psi(\bar{\theta}, w).$$

Note that $\Psi$ is identical to its definition for TRBP, but within a different domain of weights. We define $\mathcal{D} = \{w | \sum_r w_r = 1\}$ to be the domain of normalized weights, $\mathcal{D}^+$ the domain of weights in which Jensen's inequality holds, and $\mathcal{D}^-$ the domain in which the reversed Jensen's inequality holds. Note that $\mathcal{D}^+$ is a convex set (the probability simplex), while $\mathcal{D}^-$ is a union of $R$ disjoint convex sets: $\mathcal{D}^- = \cup_r \mathcal{D}_r^-$, where $\mathcal{D}_r^- = \{w | w \in \mathcal{D}, w_r > 1, w_{r'} < 0 \text{ for } r' \neq r\}$ and each subdomain $\mathcal{D}_r^-$ corresponds to a spanning tree $T_r$. See Fig. 1 for a sketch of the domains. Each $\mathcal{D}_r^-$ is a infinite convex set, but their union $\mathcal{D}^-$ is not convex.

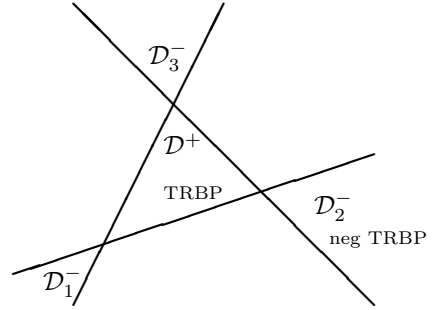

Figure 1: An illustration of the weight domains on a graph with three nodes and edges. The plane represents the domain $\mathcal{D} = \{w | \sum w_r = 1\}$, where the $w_i$ are the weights of the three possible spanning trees (see Fig. 2 for more detail). All-positive weights form the probability simplex $\mathcal{D}^+ = \{w | w \in \mathcal{D}, w_r > 0 \ \forall r\}$, while $\mathcal{D}_r^- = \{w | w \in \mathcal{D}, w_r > 0, w_{r'} < 0 \text{ for } r' \neq r\}$ are subdomains of negative weights. The negative domain is $\mathcal{D}^- = \cup_r \mathcal{D}_r^-$, a union of three disjoint infinite sets.

Again, a natural question is how to choose the optimal $\bar{\theta}$ and $w$ to obtain the tightest bound. However, for negative weights this optimization is non-convex. One obvious difficulty is the non-convexity of the domain $\mathcal{D}^-$. In fact, searching for the optimal weights requires not only a local search inside any given subdomain $\mathcal{D}_r^-$, but also global moves between the subdomains (see Fig. 1). The latter task is particularly challenging, since the number of subdomains is very large (equal to the number of possible spanning trees).

Another difficulty is that $\Psi(\bar{\theta}, w)$ is neither a convex nor concave function of $\bar{\theta}$ for $w \in \mathcal{D}^-$. To see this, one need only notice that $\Psi(\bar{\theta}, w)$ is convex in $\theta^1$ if $w_1 > 0$, and is concave in $\theta^r$ for $r = 2, \ldots, R$ if $w_r < 0$. The original TRBP algorithm derives a message-passing representation using strong duality of the upper bound, but for negative weights this argument will not hold. Instead, as with mean field we search for a local optimum, and if desired we can always perform multiple restarts and select the tightest.

## 4 TREE MATCHING CONDITIONS

To facilitate further discussion, we establish some properties of the stationary points of $\Psi(\bar{\theta}, w)$. This process proceeds similarly to the derivation of BP from the Bethe free energy given by Yedidia et al. (2004), and our resulting conditions are identical to those in Wainwright et al. (2005), derived using the convex dual function. However, we establish them directly as properties of any stationary point in order to apply them to our non-convex $\Psi(\theta, w)$ and find local maxima.

Let $p_r(x | \theta^r) = \exp(\frac{1}{w_r} \sum_\alpha \theta_\alpha^r \phi_\alpha - \Phi(\frac{\theta^r}{w_r}))$ be the dis-

tribution over the spanning tree $T_r$. We let $\tau_i^r(x_i)$ ($i \in V$) and $\tau_{ij}^r(x_i, x_j)$, $(i,j) \in E_r$, be the single and pairwise marginals of $p_r(x|\theta^r)$.

The gradient of $\Psi(\bar{\theta}, w)$ w.r.t. $\bar{\theta}$ is given by

$$\frac{\partial \Psi(\bar{\theta}, w)}{\partial \theta_\alpha^r} = \sum_x p_r(x|\theta^r) \phi_\alpha(x).$$

For the overcomplete representation, the $\phi(x)$ are indicator functions, and the gradients are the corresponding marginals of $p_r(x|\theta^r)$:

$$\frac{\partial \Psi(\bar{\theta}, w)}{\partial \theta_i^r(x_i)} = \tau_i^r(x_i) \quad \text{and} \quad \frac{\partial \Psi(\bar{\theta}, w)}{\partial \theta_{ij}^r(x_i, x_j)} = \tau_{ij}^r(x_i, x_j).$$

Adding the equality constraint $\sum_r \theta_r = \theta$ as a Lagrange multiplier, we find the conditions

$$\tau_i^r(x_i) = \tau_i(x_i) \qquad \forall r$$
$$\tau_{ij}^r(x_i, x_j) = \tau_{ij}(x_i, x_j) \qquad \forall r \in T(i,j)$$

where $\tau_i(x_i)$ and $\tau_{ij}(x_i, x_j)$ are the Lagrange multipliers. Thus at any stationary point, the single and pairwise marginals of every spanning tree are matched, with common marginals $\tau_i$, $\tau_{ij}$.

Similarly, we take the derivative of $\Psi(\bar{\theta}, w)$ w.r.t. $w$:

$$\frac{\partial \Psi(\bar{\theta}, w)}{\partial w_r} = -\sum_x p_r(x|\theta^r) \log p_r(x|\theta^r) = H_r,$$

where $H_r$ is the entropy of $p_r(x|\theta^r)$. Again adding the constraint $\sum_r w_r = 1$ using a Lagrange multiplier, the stationary condition for $w$ is

$$H_r = H \qquad \forall r,$$

where $H$ is the Lagrange mulitplier, showing that at the stationary point, the entropies of each the spanning tree should match with each other, with $H$ representing the common entropy of all the spanning trees.

## 5   MESSAGE PASSING UPDATES

Based on the insight that the stationary points of $\Psi$ satisfy marginal matching conditions, we can derive a message passing fixed point algorithm to optimize the bound. Again, our results will closely mirror the update scheme of Wainwright et al. (2005), but do not rely on strong duality and are applicable to our negative weight regime.

We begin by writing the factorization of tree $r$ as

$$p_r(x|\theta^r) \propto \prod_i \psi_{i,r}(x_i) \prod_{(i,j) \in E_r} \psi_{ij,r}(x_i, x_j).$$

Subject to the constraint $\sum_r \theta^r = \theta$ it is easy to see

$$\prod_r \psi_{i,r}^{w_r} = \psi_i(x_i) \quad \text{and} \quad \prod_{r \in T(i,j)} \psi_{ij,r}^{w_r} = \psi_{ij}.$$

Since $p_r(x|\theta^r)$ is tree-structured, its marginals can be found by forward-backward message passing; the update on $T_r$ is

$$m_{ij}^r(x_j) \propto \sum_{x_i} \psi_{i,r}(x_i) \psi_{ij,r}(x_i, x_j) \frac{m_{\sim i}^r(x_i)}{m_{ji}^r(x_i)}.$$

where $m_{\sim i}^r(x_i) = \prod_{k \in \mathcal{N}_r(i)} m_{k,i}^r(x_i)$ is the product of the messages sending to node $i$ from its neighborhood $\mathcal{N}_r(i)$ in $T_r$. The marginals of $T_r$ are

$$\tau_i^r \propto \psi_{i,r}(x_i) m_{\sim i}^r(x_i)$$
$$\tau_{ij}^r \propto \psi_{i,r}(x_i) \psi_{j,r}(x_j) \psi_{ij,r}(x_i, x_j) \frac{m_{\sim i}^r(x_i) m_{\sim j}^r(x_j)}{m_{ij}^r(x_j) m_{ji}^r(x_i)}$$

for $(i,j) \in E_r$. At any stationary point, the marginals must be matched, $\tau_i(x_i) \propto \psi_{i,r}(x_i) m_{\sim i}^r(x_i)$, so that

$$\tau_i = \prod_r \tau_i(x_i)^{w_r} = \psi_i(x_i) \prod_r [m_{\sim i}^r(x_i)]^{w_r} \quad (2)$$

We define "overall messages"

$$m_{ij} = \prod_{r \in T(i,j)} \left(m_{ij}^r\right)^{w_r} \quad \text{and} \quad m_{\sim i} = \prod_{k \in \mathcal{N}(i)} m_{ki}. \quad (3)$$

Multiplying together the pairwise marginals, we have

$$\tau_{ij}^{\mu_{ij}} \tau_i^{(1-\mu_{ij})} \tau_j^{(1-\mu_{ij})} = \prod_{r \in T(i,j)} \tau_{ij} \prod_{r \notin T(i,j)} \tau_i \tau_j$$
$$\propto \psi_i(x_i) \psi_j(x_j) \psi_{ij}(x_i, x_j) \frac{m_{\sim i}(x_i) m_{\sim j}(x_j)}{m_{ij}(x_j) m_{ji}(x_i)},$$
$$\Rightarrow \quad \tau_{ij} \propto \psi_i \psi_j \psi_{ij}^{1/\mu_{ij}} \frac{m_{\sim i} m_{\sim j}}{m_{ij}^{1/\mu_{ij}} m_{ji}^{1/\mu_{ij}}}. \quad (4)$$

Applying $\sum_i \tau_{ij}(x_i, x_j) = \tau_j(x_j)$ and (2)–(4) gives

$$m_{ij}(x_j) \propto \left[ \sum_{x_i} \psi_i(x_i) \psi_{ij}(x_i, x_j)^{1/\mu_{ij}} \frac{m_{\sim i}(x_i)}{m_{ji}(x_i)^{1/\mu_{ij}}} \right]^{\mu_{ij}} \quad (5)$$

Substituting $\bar{m}_{ij} = m_{ij}^{1/\mu_{ij}}$ yields the same message passing algorithm as that in Wainwright et al. (2005). We use the form (5) as it does not diverge even for $\mu_{ij} \to \pm 0$ or $\mu_{ij} \to \pm \infty$, and is also convenient to draw connections to mean field in Section 7.

Finally, since $p_r(x|\theta^r)$ is a tree-structured distribution with marginals $\tau_i$ and $\tau_{ij}$, its log-partition function

can be written as a sum of single-node entropies and pairwise mutual informations:

$$\Phi\left(\frac{\theta_r}{w_r}\right) = \frac{1}{w_r}\langle\tau, \theta^r\rangle + \sum_{i \in V} H_i(\tau) - \sum_{(i,j) \in E_r} I_{ij}(\tau),$$

where $H_i(\tau) = -\sum_s \tau_{i;s} \log \tau_{i;s}$ is the single-node entropy, and $I_{ij}(\tau) = \sum_{s,t} \tau_{ij;st} \log \frac{\tau_{ij;st}}{\tau_{i;s}\tau_{j;t}}$ is the mutual information. The corresponding upper or lower bound at a stationary point with weights $w$ is

$$\Psi_w = \sum_r w_r \Phi\left(\frac{\theta^r}{w_r}\right)$$
$$= \langle\tau, \theta\rangle + \sum_{i \in V} H_i - \sum_{(i,j) \in E} \mu_{ij} I_{ij} \triangleq -F(\tau, \mu). \quad (6)$$

$F(\tau, \mu)$ is introduced in Wainwright et al. (2005) as the dual function of the upper bound and interpreted as a free energy. Here we derive it directly, and do not require convexity as is necessary for strong duality. Since the fixed point of the message passing algorithm (5) is a stationary point of $\Psi(\bar{\theta}, w)$, (5) can be used to search for local maxima even when $\Psi(\bar{\theta}, w)$ is non-convex. Conversely, Wainwright et al. (2005) shows that the fixed point with respect to $\tau_i$ and $\tau_{ij}$ of the free energy (6) satisfies the message update equation (5). Thus, message passing via (5) corresponds to the stationary points of both $\Psi(\bar{\theta}, w)$ and $F(\tau, \mu)$. We use these two points of view interchangably in the sequel. It is worth emphasizing that $\Psi(\bar{\theta}, w)$ and $-F(\tau, \mu)$ only coincide at stationary points, and can be very different elsewhere. For example, $\Psi(\bar{\theta}, w)$ is always upper bounded by $\Phi(\theta)$ but $-F(\tau, \mu)$ with negative weights has no upper bound, and local maxima of $\Psi(\bar{\theta}, w)$ can be saddle points of $-F(\tau, \mu)$ (see Section 7).

## 6 OPTIMIZING OVER WEIGHTS

Wainwright et al. (2005) optimizes the (positive) weights using conditional gradient descent, which they show is equivalent to solving a maximum-weight spanning tree problem at each iteration. Unfortunately, this method is not immediately applicable to negative weights, since the domain $\mathcal{D}^-$ is not convex. A search strategy on $\mathcal{D}^-$ requires a combination of local and global optimization. By choosing a representation carefully, we are able to optimize over the positive and negative trees separately, and use the same optimization techniques of TRBP on the negative weights.

Let $T^+ = T_{r_0}$ be the spanning tree with positive weight $w^+ = \beta + 1$, where $\beta > 0$, and $\mathcal{T}^- = \{T_r^-\}$ be a collection of spanning trees with negative weights $w^- = \{-\beta v_r\}$, where $\sum_r v_r = 1$ and $0 \leq v_r \leq 1$. Let $H^+$ be the entropy of $T^+$ and $H_r^-$ the entropy of $T_r^-$. For convenience, we also include $T^+$ as a negative tree, so that the feasible range of $v_r$ is in fact $\mathcal{D}^+$, the same as that in regular TRBP. It is then easy to see that

$$\frac{\partial \Psi}{\partial \beta} = H^+ - \sum_r v_r H_r^- \quad \text{and} \quad \frac{\partial \Psi}{\partial v_r} = -\beta H_r^-.$$

For fixed $T^+$, the optimization of $\beta$ and $v_r$ corresponds to a local search inside each subdomain, and can be updated using the gradient. More formally, let $\beta^{(n)}$ and $v_r^{(n)}$ be the values at iteration $n$. Because we require $\beta > 0$, we use log-gradient ascent to update $\beta$: defining $\gamma = \log \beta$, we update $\gamma$ with stepsize $\epsilon_\beta > 0$, producing a corresponding update rule for $\beta$:

$$\beta^{(n+1)} = \beta^{(n)} \exp(\epsilon_\beta (H^+ - \sum_r v_r H_r^-) \beta^{(n)}). \quad (7)$$

The negative weights $v = \{v_r\}$ can be updated via conditional gradient:

$$\tilde{v}_r^{(n+1)} = \arg \max_{v \in \mathcal{D}^+} \{-\sum_r H_r^- (v_r - v_r^{(n)})\}$$
$$v^{(n+1)} = v^n + \epsilon_v(\tilde{v}^{(n+1)} - v^{(n)}) \quad (8)$$

As in Wainwright et al. (2005), $H_r^-$ can be written as a sum of entropy and mutual information terms,

$$H_r^- = \sum_{i \in V} H_i^-(\tau) - \sum_{(i,j) \in E_r^-} I_{ij}^-(\tau);$$

the first term is constant for all $r$, and computing $\tilde{v}^{(n+1)}$ is solved as a maximal spanning tree problem using mutual information as the edge weights.

Choosing the positive tree $T^+$ now corresponds to "jumping" between the different subdomains of $\mathcal{D}^-$, a combinatorial optimization problem for which the local gradient may not be helpful. Since the number of subdomains of $\mathcal{D}^-$ equals the (possibly superexponential) number of spanning trees, it is challenging to find an optimal $T^+$. Instead, we employ a heuristic for re-selecting $T^+$ that appears to work well in practice. Analysis connecting our method to mean field shows that $T^+$ can be interpreted as the skeleton of $G$, along which the original variable dependencies are more closely maintained. From this viewpoint, it makes sense to choose $T^+$ to match the distribution as closely as possible, and an obvious heuristic to use is again to employ the Chow-Liu algorithm (Chow and Liu, 1968) – we choose $T^+$ via a maximum spanning tree problem in which the mutual information of the previous iteration serves as the weight on each edge:

$$T^+ = \arg\max\{\sum_{(i,j) \in E^+} I_{ij}\}. \quad (9)$$

Our experiments (Section 8) suggest that the updates Eqn. (9) can significantly improve the lower bound.

# 7 OTHER RELATED WORK

Structured mean field is a variational inference algorithm that approximates the original graphical model using a tractable subgraph. It is a extension of naïve mean field, in which the tractable subgraph is fully disconnected. Bouchard-Cote and Jordan (2009) show that the computational complexity of optimizing the structured mean field objective depends on the type of tractable subgraph chosen. Formally, they define:

**Definition 7.1.** *An acyclic subgraph with edges $E' \subseteq E$ is called v-acyclic (very acyclic) if for all $(i,j) \in E$, $E' \cup (i,j)$ is still acyclic. Otherwise, it is called b-acyclic (barely acyclic).*

Bouchard-Cote and Jordan (2009) go on to show that structured mean field with a *v*-acyclic subgraph is easier to solve, compared to a *b*-acyclic subgraph. We now show that structured mean field using a *v*-acyclic subgraph is a special case of our algorithm.

**Theorem 7.1.** *Let $T_v$ be a v-acyclic subgraph of the original graph $G$. Let us construct a negative TRBP approximation by taking positive tree $T^+ = T_v$ with weight $\beta + 1$, and negative trees which all include $T^+$ as a subtree, $\mathcal{T}^- = \{T_v \cup (i,j) : (i,j) \in E \setminus E_v\}$, with weights $\{-\beta v_r\}$. Then when $\beta \to +\infty$, negative TRBP is equivalent to structured mean field with $T_v$.*

*Proof.* We first note that by the definition of a *v*-acyclic subgraph, $\mathcal{T}^-$ is a collection of subtrees. As $\beta \to \infty$, the edge appearance probabilities $\mu_{ij} \to -\infty$ for all edges not in $T_v$, and $\mu_{ij} = 1$ for edges in $T_v$ (as every negative tree also includes them). Now consider the message passing algorithm (5) as the stationary point of the free energy (6), when $\mu_{ij}$ approaches infinity. The derivative of $F(\tau, \mu)$ w.r.t. $\tau_{ij}$ is

$$\frac{\partial F(\tau, \mu)}{\partial \tau_{ij}(x_i, x_j)} \approx -\mu_{ij}\left(\log \frac{\tau_{ij}(x_i, x_j)}{\tau_i(x_i)\tau_j(x_j)}) - 1\right)$$

which equals zero iff $\tau_{ij} = \tau_i \tau_j$. Thus the mutual information term $I_{ij} = 0$ for the corresponding edge $(i,j) \in E \setminus E_v$. It then is easy to show that $F(\tau, \mu)$ exactly corresponds to the free energy minimized by structured mean field (Bouchard-Cote and Jordan, 2009, Equation (6)). □

A consequence of Theorem 7.1 is that naïve mean field is a special case of our algorithm with $T^+ = (G, \emptyset)$ and $\beta \to +\infty$, in which case all the edge appearance probabilities are infinite, i.e., $\mu_{ij} \to -\infty$ for $(i,j) \in E$. It is also straightforward to directly derive naïve mean field updates as a special case of algorithm (5) as $\mu_{ij} \to -\infty$, by applying the fact that the geometric mean is the limit of the power mean as the power (in this case, $\frac{1}{\mu}$) approaches zero.

Interestingly, loopy BP can also be viewed as negative tree weights. Wainwright et al. (2005) notes that the Bethe free energy is obtained from (6) when all $\mu_{ij} = 1$. While not achieved in either $\mathcal{D}^+$ or $\mathcal{D}^-$, this can be achieved using weights that includes multiple positive weights and some negative weights; see Fig. 2 for an illustrative example. Notionally, loopy BP can be thought of as corresponding to some subspace of $\mathcal{D}$ in Fig. 1 between the negative sub-domains (although technically it requires non-spanning trees and thus cannot be visualized in only the two dimensions pictured). This view reinforces why loopy BP guarantees neither an upper nor lower bound in general.

Our method is also closely related to the framework of fractional belief propagation (Wiegerinck and Heskes, 2003). While the fractional BP framework also highlights connections to mean field and TRBP, it does not give intuition on how to choose the weights so as to guarantee upper or lower bounds on the log partition function, nor does it consider the use of negative valued weights or the connections to structured mean field. Our framework clearly delineates the regimes in which upper and lower bounds are guaranteed, extending and connecting these algorithms more precisely.

Another way of relating TRBP and mean field is through the variational dual viewpoint (Wainwright and Jordan, 2008), which writes the log-partition $\Phi$ as the supremum of the negative free energy over the marginal polytope $\mathcal{M}$. TRBP approximates $\mathcal{M}$ by a convex outer bound $\mathcal{M}_T$, and replaces the free energy by a convex lower bound, yielding an upper bound of $\Phi$; mean field minimizes the true free energy over an non-convex inner bound $\mathcal{M}_{MF} \subseteq \mathcal{M}$, yielding a lower bound. As shown in Section 5, negative TRBP instead finds a stationary point of a non-convex approximate free energy (6) over the outer bound $\mathcal{M}_T$, which perhaps surprisingly still yields a lower bound on $\Phi$. This inherent non-convexity makes it difficult to understand negative TRBP directly from the dual polytope view.

Additionally, we note that the stationary points of negative TRBP often correspond to a saddle point of the approximate free energy. To illustrate this point, consider the case when $T^+ = (V, \emptyset)$, making all $\mu_{ij}$ negative. The free energy (6) is then

$$-F(\tau, \mu) = \langle \tau, \theta \rangle + \sum_{i \in V} H_i - \sum_{(i,j) \in E} \mu_{ij} I_{ij}.$$

As $\mu_{ij} \to -\infty$, the stationary point $\tau^*$ of negative TRBP approaches a fixed point of mean field, and hence is a local minimum of $F(\tau, \mu)$ in $\mathcal{M}_{MF}$ (in which the marginals satisfy $\tau_{ij} = \tau_i \tau_j$). In contrast, outside $\mathcal{M}_{MF}$ we have $I_{ij} \geq 0$, yielding $F(\tau, \mu) \approx \sum \mu_{ij} I_{ij} \to -\infty$, and hence $\tau^*$ is a local maximum of $F(\tau, \mu)$ in this subspace. Thus, $\tau^*$ is a saddle point.

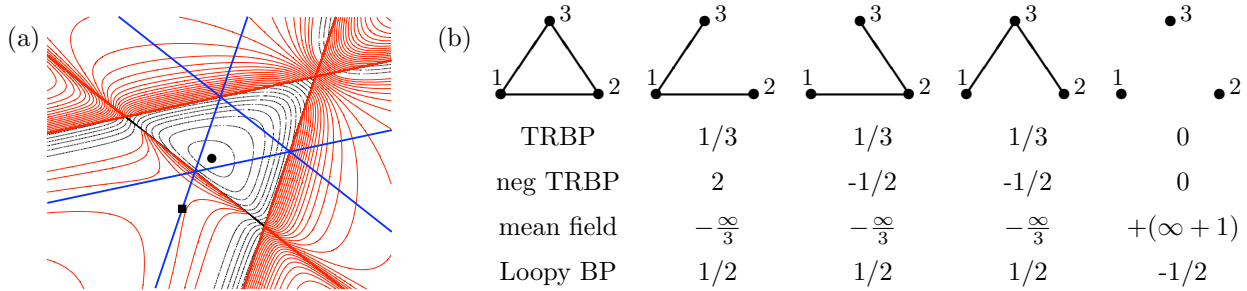

Figure 2: (a) The contour of $\Psi_w$ in a triangle graph (parameters given in the text). The black contours indicate regions in which $\Psi_w$ is larger than the true log partition function $\Phi(\theta)$, and red lines regions in which $\Psi_w$ is smaller than $\Phi(\theta)$. The circle denotes the optimal weight of TRBP, and the square denotes the optimal weights of negative TRBP. (b) An illustration of how different algorithms are understood as a combination of subtrees with different weights. The original graph (top left) is 3 nodes and fully connected; there are 3 spanning trees and we include one fully disconnected subgraph listed to its right. In all cases, the subtree weights sum to one.

## 8 EXAMPLES AND EXPERIMENTS

We first illustrate the properties of positive and negative TRBP on a small example in which the space of approximations can be directly visualized. We then show the algorithm's behavior and improvement over mean field on attractive and mixed-sign Ising models.

### 8.1 TRIANGLE GRAPH

To visualize the algorithm's bounds, we examine a simple triangle graph $G$ in Fig. 2b. $G$ has three spanning trees $T_r$, $r = 1 \ldots 3$, also listed in the figure. To make the connection to naïve mean field and loopy belief propagation, we also include a non-spanning tree (the disconnected graph). The parameters of the original distribution are $\psi_i = [1, 1]^T$ for $i = 1, 2, 3$ and

$$\psi_{12} = \begin{bmatrix} 1 & 0.8 \\ 0.8 & 1 \end{bmatrix} \quad \text{and} \quad \psi_{13} = \psi_{23} = \begin{bmatrix} 1 & 0.5 \\ 0.5 & 1 \end{bmatrix}.$$

Different algorithms then correspond to different regions of the weights; the table in Fig. 2b gives some illustrative examples of valid weight values for each algorithm. The general concept of the relationship between these domains is also illustrated in Fig. 1, although again it does not include the dimension of non-spanning tree weight required to express the domain of mean field or loopy BP.

For this small example, we can illustrate the behavior of the bound for values of the $w_r$ (now using only the spanning trees). For each fixed value of $w \in \mathcal{D}$, we use the message passing algorithm to optimize (finding a stationary point of) $\Psi(\bar{\theta}, w)$, and denote this quantity $\Psi_w$. We plot the contour of $\Psi_w$ in Fig. 2, indicating contours above the true partition function in black and those below in red (best viewed in color).

As can be seen, the positive domain $\mathcal{D}^+$ guarantees an upper bound and the negative domain $\mathcal{D}^-$ guarantees a lower bound, consequences of Jensen's and the reversed Jensen's inequality. The other regions guarantee neither an upper nor a lower bound. Notice that although $\Psi_w$ is convex in $\mathcal{D}^+$, it is generally non-convex and even discontinuous outside $\mathcal{D}^+$, with saddle points and singularities visible in the figure.

The optimal weights of TRBP and negative TRBP are labeled with a black circle and a square, respectively. Notice in particular that the optimal point of negative TRBP lies near the interior in this example, with the bound decreasing outward. Since the infinity point corresponds to mean field, this shows a simple concrete example in which negative TRBP improves on the mean field bound.

### 8.2 ISING MODELS

We also verify our method using simulations on 2D, $10 \times 10$ Ising models (on which the exact partition function can still be computed tractably). Let $G = (V, E)$ be a square lattice, with nodes corresponding to binary variables $x = \{x_i\}$, $x_i \in \{-1, 1\}$, and

$$p(x) = \exp\left( \sum_{i \in V} \theta_i x_i + \sum_{(i,j) \in E} \theta_{ij} x_i x_j - \Phi(\theta) \right).$$

We follow Wainwright et al. (2005) and generate the parameter vector $\theta$ randomly. We draw $\theta_i \sim \mathcal{U}(-0.05, 0.05)$ for each node $i \in V$, and for each edge $(i, j) \in E$, we sample $\theta_{ij}$ independently using either

1) for attractive interactions, $\theta_{ij} \sim \mathcal{U}(0, c)$;
2) for mixed interactions, $\theta_{ij} \sim \mathcal{U}(-c, c)$.

We vary $c$ from 0 to 2, and perform 20 simulations at each value. Results are shown relative to the exact partition function, computed by junction tree. Lower bounds are calculated using both negative TRBP and naïve mean field; for comparison, we also calculate the upper bound found using TRBP. For mean field, we re-run the algorithm starting from uniform marginals and ten random initializations, and select the best lower bound. Negative TRBP is run only once, with initial $\beta = 10$, a randomly selected positive tree, and $\mu_{ij}$ created by drawing spanning trees until all edges are covered. Step sizes used were $\epsilon_\beta = 1$ and $\epsilon_v = 0.05$.

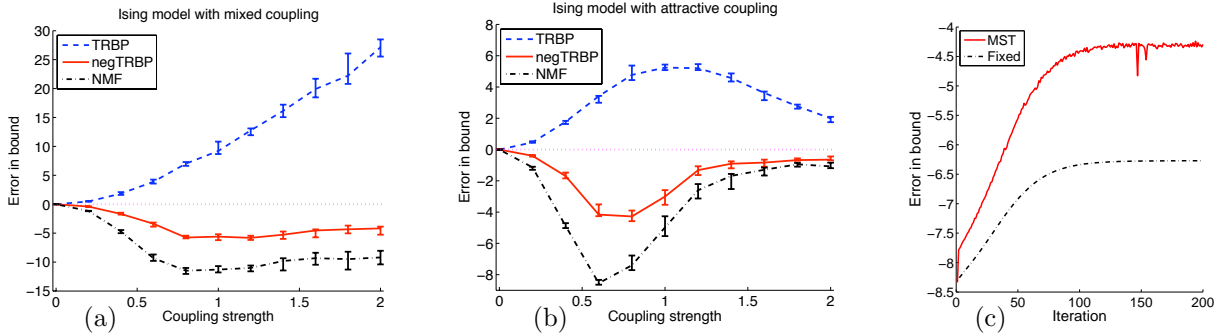

Figure 3: (a)-(b) Comparisons of TRBP, negative TRBP and naive mean field on the Ising model with (a) mixed coupling strength and (b) attractive coupling strength. The curves represent the median error over 20 trials, and the error bars are 25% and 75% percentiles. (c) Lower bound found by negative TRBP as a function of iteration for a fixed positive tree, and when the positive tree is updated using the Chow-Liu heuristic. Updating the choice of positive tree yields a much tighter lower bound.

Fig. 3a–b show the results of these experiments (median values, with interquartile range shown as error bars). Negative TRBP gives a significant improvement over naïve mean field (typically about a factor of two) on most of the domain.

To see the effect of our choice of positive spanning tree, we compare the bound obtained using a fixed positive spanning tree to that obtained using our Chow-Liu heuristic, as a function of iteration on an attractive model with $c = .6$; the results are shown in Fig. 3c. Although in some domains there may be a natural structure to emphasize, one advantage of our approach is its ability to adaptively search for the best structure during inference, as the importance of dependencies becomes evident. This can be seen in the figure, as adapting the structure greatly tightens the bound.

## 9 CONCLUSION

In this paper, we have established a reversed Jensen's inequality and applied it to develop a lower bound of the log partition function. Our algorithm mirrors the update equations of TRBP, showing that upper and lower bounds are obtained via the same algorithm under different weight conditions. We also have shown that structured mean field on any $v$-acyclic subgraph is a limiting case of our algorithm. Our algorithm improves significantly on naïve mean field, and furthermore can use its current results to modify the structure of the underlying approximation, improving on bounds using a single fixed structure. Open questions include improving the optimization, particularly the combinatorial choice of positive tree, and whether the bounds can be further tightened, such as exploring the relationship of $b$-acyclic structured mean field.

**Acknowledgements:** This work was supported in part by NIH grants P50-GM076516 and NIAMS AR 44882 BIRT revision.